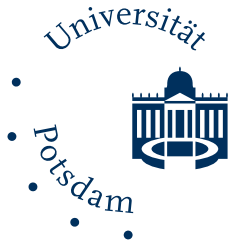 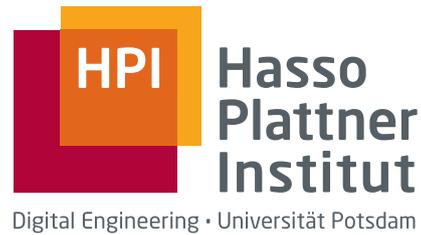

# Evaluation of Ensemble Learning Techniques for handwritten OCR Improvement

Evaluierung von Ensemble Lern Techniken zur Verbesserung von hangeschriebener OCR

## Martin Preiß

Universitätsbachelorarbeit
zur Erlangung des akademischen Grades

### Bachelor of Science
*(B. Sc.)*

im Studiengang
IT Systems Engineering

eingereicht am 29. July 2022 am
Fachgebiet Digital Health & Machine Learning der
Digital-Engineering-Fakultät
der Universität Potsdam

**Gutachter**     Prof. Dr. Christoph Lippert
**Betreuer**     Benjamin Bergner

# Abstract

For the bachelor project 2021 of Professor Lippert's research group, handwritten entries of historical patient records needed to be digitized using Optical Character Recognition (OCR) methods. Since the data will be used in the future, a high degree of accuracy is naturally required. Especially in the medical field this has even more importance. Ensemble Learning is a method that combines several machine learning methods. This procedure is claimed to be able to achieve an increased accuracy for existing methods. For this reason, Ensemble Learning in combination with OCR is investigated in this work in order to create added value for the digitization of the patient records. It was possible to discover that ensemble learning can lead to an increased accuracy for OCR, which methods were able to achieve this and that the size of the training data set did not play a role here.

# Zusammenfassung

Für das Bachelorprojekt 2021 des Lehrstuhls von Professor Lippert sollten handgeschriebene Einträge von historischen Patientenakten mithilfe von Optical Character Recognition (OCR) Methoden digitalisiert werden. Da die Daten weiterhin verwendet werden sollen, ist natürlich eine hohe Genauigkeit erforderlich. Gerade im medizinischen Bereich gewinnt dies noch mehr an Bedeutung. Ensemble Learning ist ein Verfahren welches mehrere Machine Learning Methoden kombiniert. Diesem Verfahren wird die Fähigkeit nachgesagt, für bestehenden Mehtoden eine erhöhte Genauigkeiten dieser erreichen zu können. Aus diesem Grund wird Ensemble Learning in Kombination mit OCR in dieser Arbeit untersucht, um für die Digatlisierung der Patientenakten einen Mehrwert zu schaffen. Dabei konnte beantwortet werden, dass Ensemble Learning für OCR zu einer erhöhten Accuracy führen kann, welche Methoden dies umsetzten und dass die größte des Trainingsdatensatzes hier keine Rolle spielte.

# Acknowledgments

Since all the people I want to thank are fluent in german, I have written the acknowledgment in german. Nevertheless an english translation can be found below.

**German original**   Ich habe sehr stark dafür kämpfen müssen um am HPI studieren zu dürfen. Deshalb möchte ich als erstes den Menschen in meinem Leben danken die es mir überhaupt erst ermöglicht haben diese Zeilen zu schreiben. Mein Dank geht an meine Familie die mich jetzt nun 23 Jahre lang auf diesem Planeten unterstützt. Danke Papa, dass du stets so darauf erpicht warst, dass aus mir was wird. Danke Mama, dass du stets drauf geachtet hast, dass ich auch ein gesunder Bub werde. Danke Schwesterherz, dass ich immer auf dich zählen kann. Ich bin sehr stolz auf dich. Natürlich will ich auch nicht meine Omas, Tanten, Patentante, Cousinen, Onkel und Freunde vergessen. Ohne euch wäre ich nicht wo ich heute stehe. Vielen Danke für eure Unterstützung und Zuspruch all die Jahre.

Auch möchte ich den Betreuern des Bachelorprojekts Benjamin Bergner und Prof. Dr. Christoph Lippert für die Unterstützung das letze Jahr danken. Ich hab das Jahr als sehr spannend empfunden und freue mich, dass wir ein wenig Wissen von Ihnen abschöpfen durften.

Zum Schluss möchte ich noch meinem Team danken, welches mich das letzte Jahr begleiten durfte. Wir waren klasse Leute und jeder von euch hat seinen eigenen Satz verdient . Was ihr mit diesem macht ist eure Sache (zwinkersmiley). Danke Cedric, für deinen starken unterstützenden Rücken. Danke Smilla, für ihre Güte uns in den letzten Monaten zu behüten. Danke Elena, die mir half TrOCR zu bezwingen. Danke Paul, dem zweit besten Datenputzer dieses Teams :P. Danke Romeos Gehirn. Danke Tom , dass er nicht wahnsinnig durch mich geworden ist . Danke Abdu, der mir immer half neues zu lernen. Natürlich danke ich auch dem Leser dieser Arbeit, dass er sich die Zeit nimmt in meinen Zeilen zu verweilen. Ich wünsche viel Spaß beim Lesen.

**English translation**   I had to fight very hard to be able to study at HPI. Therefore, I would first like to thank the people in my life who made it possible for me to write these lines. My thanks goes to my family who has supported me on this planet for 23 years now. Thank you dad for always being so eager with me to become successful. Thank you mom for always making sure that I would be a healthy lad. Thank you sister, that I can always count on you. I am very proud of you. Of course I don't want to forget my grandmas, aunts, godmother, cousins, uncles and friends. Without you I would not be where I am today. Thank you for your support and encouragement all these years.

I would also like to thank the supervisors of the bachelor project Benjamin Bergner and Prof. Dr. Christoph Lippert for their support during the last year. I found the year very inspiring and I am glad that we were able to gain some knowledge from them. Finally, I would like to thank my team, which was allowed to accompany me during the last year. We have been great folks and each of you deserves your own sentence. What you do with it is up to you (wink wink). Thank you Cedric, for your strong supportive back. Thank you Smilla, for your kindness to shepherd us in the last months. Thank you Elena, for helping me defeat TrOCR. Thank you Paul, the second best data cleaner of this team :P. Thank you Romeo's brain. Thanks Tom for not going insane because of me . Thanks Abdu who always helped me to learn new things. Of course I also thank the reader of this work for taking the time to dwell in my lines. I wish a lot of fun while reading.

# Contents

# List of Figures

# List of Tables

# 1 Introduction

## 1.1 Motivation

As part of the 2021 bachelor's project "Human in the Loop: Deep Learning on handwritten Patient Records" of Professor Lippert's research group, historical patient records of the Walter Kempner Diet had to be digitized for the Duke University. For this purpose, various modern Optical Character Recognition (OCR) methods were used to extract the handwriting contained in the documents. This provided acceptable results. But since the digitized data will be used for medical studies, it is important that they are captured as accurately as possible. After all, important decisions and findings will be made later on the basis of this data. During the practical implementation of the OCR methods for the bachelor project, the idea of using several of these methods together came up independently. The methods had different accuracies and the idea of combining them was quickly considered. On the one hand, there would be no need to commit to only one method and discard the others. On the other hand, it was also questioned whether a combination could lead to an increased accuracy. After a short research, it was discovered that the idea is not new and is summarized under the term Ensemble Learning. Since many blog articles on this topic also promised increased accuracy [1, 2], it was decided to get to the bottom of this topic in combination with OCR in order to hopefully create added value for the digitization of patient records.

## 1.2 Contribution

For this purpose, the following research questions will be answered in this paper. First, the main question is whether Ensemble Learning can really improve OCR. After all, further research questions on this topic would make less sense if Ensemble Learning is useless for OCR anyway. This is formulated in **RQ1** as follows: "Can Ensemble Learning improve the Accuracy of modern OCR methods?" If the answer to RQ1 is yes, it is of course also of particular interest which methods have created this added benefit. The reason for this is that ensemble learning is a large topic and several approaches can be pursued here. Therefore, the goal of **RQ2** is to determine "Which Ensemble Learning methods are the most valuable?" One of the main difficulties of the bachelor project, not mentioned so far, was the relatively small data set, which was needed for the implementation of OCR. Therefore, the last question to be asked in **RQ3** is: "Can Ensemble Learning add significantly better value on smaller datasets?". Since it would be very interesting to know, if Ensemble Learning can help more on small datasets compared to large datasets. To answer these 3 research questions, this paper is structured as follows. First of all, in the background section, important terms around OCR and Ensemble Learning will be clarified for readers with less previous knowledge. Then, the methodology chapter will explain how the research questions will be approached. In the same chapter, the involved OCR methods and ensemble learning methods are named and explained. In the experiment chapter it is then described how these methods were put into practice and what results they had. These results are then evaluated in the discussion chapter to answer the research questions. The statements presented there are then summarized at the end in the conclusion chapter and further interesting research possibilities for OCR in combination with ensemble learning are mentioned.

# 2 Background

Initially, the purpose of this section is to formally clarify important terms related to OCR and Ensemble Learning. Readers with the necessary knowledge can skip this section.

## 2.1 OCR

### 2.1.1 Definition

Optical Character Recognition (OCR) in general refers to the translation of optically available handwritten or printed characters/words/numbers (any kind of symbols) into digital form[3]. The term Handwritten Text Recognition (HTR) is often used in this context. HTR is a specialization of OCR that converts sequences of handwritten symbols (words/phrases)[4]. Although HTR would often be formally the more accurate term, in the majority of scientific papers the term OCR is used rather than HTR. Nevertheless, it is necessary to mention HTR as a synonym at this point, since the term may help in researching further work. In the further course of this work, OCR will also be used as a term for HTR.

### 2.1.2 OCR in Deep Learning

There are many possibilities to implement OCR in practice. In recent years, however, Deep Learning approaches mainly have been used[5]. Deep learning is a type of machine learning that is characterized by using multiple processing layers, each with a large number of parameters, to make a prediction [6]. Deep refers here to the large number of parameters and layers that are used by deep learning models. In the field of deep learning, OCR can be classified as a sequence labeling task for text or a classification task for individual symbols.[6, 7] Therefore, a model that implements OCR receives a sequence as input (the input image) and outputs a string (sequence labeling) or only a single character (classification). Usually, sequence labeling models are supervised learning algorithms. Since all models of this work can be classified in this category, this concept and their general procedure shall be clarified. Supervised learning refers to methods that make a label prediction on a given set of input features by prior parameter adaptation of the prediction model.

The parameter adaptation is achieved by a learning algorithm. This algorithm receives already labeled input features as input and outputs the finished model with adjusted parameters in the end. The process of "learning" is generally referred to as training. In 2.1 you can see an overview of supervised learning.

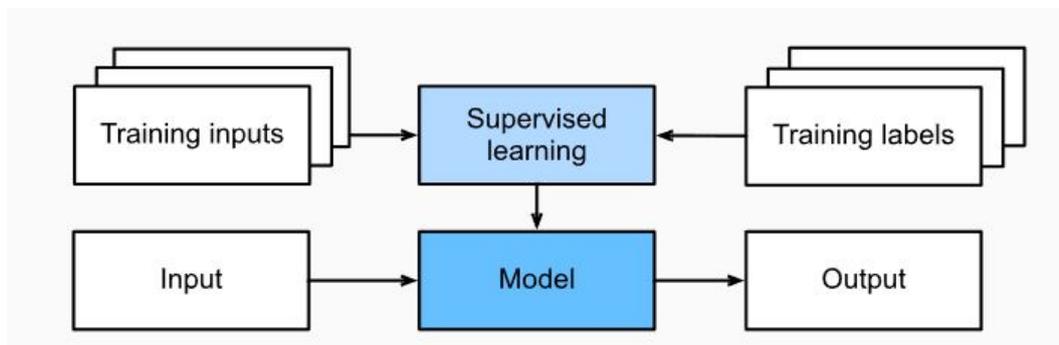

**Figure 2.1:** General supervised learning procedure [6]

## 2.2 Ensemble Learning

### 2.2.1 Definition

Ensemble Learning is a method of machine learning[8]. Usually, exactly one model is involved in making predictions, and its predictions are used directly. The difference in Ensemble Learning is that several different models are applied to generate predictions. These models are called base learners, base classifiers, or inducers [8, 9, 10]. For a given problem, all base learners perform their own prediction. The predictions of all base learners are then passed to an output component. This output component then uses all predictions to compute the final prediction. This computation can be done in a variety of ways and the most important and well-known are part of this work

### 2.2.2 Assets and Drawbacks

The main idea of Ensemble Learning is that multiple opinions combined can be more accurate than a single opinion. This basic idea was already proven in 1785 by Marquis de Condorcet in his Jury Theorem [11]. To achieve more precise estimation in ensemble learning, the following 2 principles must be satisfied [10]. First, the base learners must have as much diversity as possible in their estimates. This ensures a high variety of opinions, which can be taken into account in the final combination. On the other hand, the base learners should estimate as precisely as possible despite the diversity principle. This is important because a high diversity of opinions does not add value if all opinions are wrong. In combination, these principles allow increased accuracy in estimation as well as other benefits at statistical, computational, and representational levels [9, 10, 12]. Statistically, combining multiple models reduces the likelihood of using a single model that overfits on the training data. Of course, an ensemble can also overfit, but then the principle of diversity of base learners would be violated. In addition, individual models can remain in a local optimum in the training process. The combination of those models would allow a better approximation of the best possible optimum in the function space. The ensemble of several models also increases the probability of better generalization. In other words, to react better to new data not available in the data set, since the combination of the individual models expands the function space. Naturally, Ensemble Learning also has disadvantages. These are mainly the high costs of resources, for example time, power or memory, which are caused by the training of the base learner or the prediction of the ensemble. However, if these disadvantages can be ignored or if better accuracy is simply more important, as it is the case in this work, then ensemble learning offers itself in theory as a possible tool.

### 2.2.3 Design Levels

The construction of an ensemble and the ensemble methods that are involved with it can be practically classified on 3 levels[13]. First, at the dataset level, it has to be chosen which base learner gets which training dataset with which features. Then, at the base learner level, it must be decided what methods will be used for the base learners and of course how many base learners there are in total. Finally, at the output level, an approach for merging the individual base learner estimates must be selected. Only if the right decisions are made at these 3 levels the benefits of Ensemble Learning can be practically implemented. The Ensemble methods used for the implementation can, however, be classified on several levels.

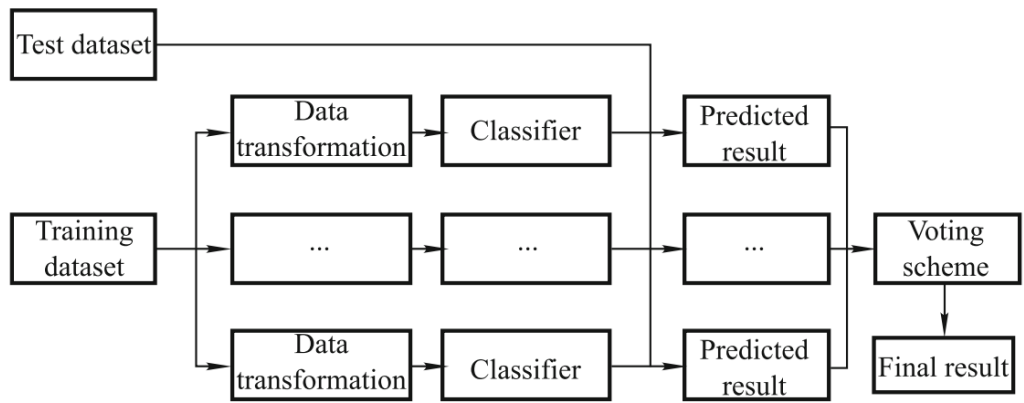

**Figure 2.2:** General structure of an Ensemble [10]

# 3 Related Work

The idea of combining ensemble learning with OCR is not new. First papers were already published in the 1990s. In 1992 for example neural networks (NN) were combined in an ensemble with the goal of recognizing handwritten digits. This ensemble achieved 20-25% better results than the best individual NN.[14] In the following years, research was further advanced and the first ensemble methods such as bagging and boosting in combination with NNs were evaluated.[15, 16, 17, 18]. Again, better accuracy results could be achieved.

Over the years, ensemble learning methods have been repeatedly evaluated in combination with other machine learning architectures such as Hidden Markov Models [12, 19, 20] or Decision Trees[21] as well as with multi-architecture ensembles [22]. Here as well, better accuracy was achieved by using ensemble methods.

However, with the beginning of the Deep Learning era, these evaluations related to OCR declined. One of the last papers known to the author is the one from 2014 [23], which deals among other topics with ensemble methods for improved recognition of historical prints. The paper evaluates some ensemble methods but this is not the main focus of the paper.

The reasons for the decline of evaluations are on the one hand the time and computational costs of training deep learning networks [24], and on the other hand the very good generic results of deep learning architectures in OCR [25]. Therefore, in recent years, there has been more research on developing single Deep Learning architectures, which already achieved excellent results even without Ensemble Learning. A good example is the paper by Ahlawat et al. [26], which aims to outperform Ensemble Learning with a single OCR model to avoid the drawbacks of ensembles.

This is certainly no reason to believe that the combination of ensemble learning with OCR is no longer relevant for research. In recent years, papers still have been published in which individual ensemble learning methods have been combined with specific OCR use cases in order to improve the results. For example, ensembles have been used to recognize handwritten persian numerals [27], farsi numerals [28], bangla letters [29], arabic words [30], musical notes [31], medieval prints/manuscripts [32, 33, 34], or bio-collections [35].

Another work worth mentioning here is the one of Matos Silva [36]. In his research, table cells from medical records were digitized using a custom-built ensemble of neural networks, which reaffirms the potential of this work in the medical context as well.

These examples show that ensemble learning is still relevant for OCR. Nevertheless, the most recent papers only refer to one or a few ensemble methods. Thus, there is a lack of an up-to-date assessment of the most important ensemble methods with respect to OCR. Outside the subject-specific OCR domain, these already exist Surveys are still published that summarize the current state of research in general [10, 37], also with respect to Deep Learning [9, 24]. In these surveys, however, OCR is mentioned at most as a possible application example.

To the best of the author's knowledge, there is a lack of studies evaluating modern OCR deep learning models in combination with ensemble learning. This work tries to fill this gap by reevaluating the most popular and important ensemble learning methods with regard to OCR.

# 4 Methodology

## 4.1 Experimental Process

To answer the research questions, the following procedure is applied. First, the state of the art OCR methods and ensemble methods are identified. Then, in the experiments, each OCR method is first evaluated on its own. Then, the ensemble methods are further analyzed. The values recorded there are then compared with the previously recorded values of the stand-alone OCR methods. This will make it possible to answer **RQ1**, whether ensemble methods can contribute to an improved accuracy with the OCR methods used. To clarify **RQ2**, the evaluation results of the ensemble methods will then be compared in order to be able to name the most effective methods. To answer **RQ3**, it is necessary to consider two data sets separately in the evaluations. With the comparison of measured values of a large and a small dataset, it can be clarified whether Ensemble Learning can add significantly better value on smaller datasets.

## 4.2 OCR

In the following, the OCR models used in this work will be named and briefly explained. For a detailed description of the methods and the exact training procedures, please visit the corresponding papers.

### 4.2.1 TrOCR

The first model to be looked at is TrOCR. TrOCR was released in 2021 by a Microsoft research team and is an end-to-end transformer-based OCR model [38]. Basically, TrOCR corresponds to a vanilla transformer encoder-decoder structure and the architecture (see figure) is structured as follows. First, TrOCR resizes the input image to a size of 384 × 384 pixels and then splits it into 16*16 sized disjoint patches. Patch embedding as well as positional embedding is then applied to the patches to retain important contextual information about the other patches. The now processed sequence of patches is then given as input to the encoder, which consists of multi-head attention modules and feed forward networks. The encoder extracts features from each of the patches, which then serves as input to the decoder. The decoder also consists of multi-head attention modules, feed forward networks, and additionally masked multi-head attention modules. With the output of the encoder, the decoder creates a probability matrix that assigns token probabilities to specific subsections of the input image. In the context domain of TrOCR, tokens refer to particular character sequences, of which TrOCR has 50265 by default. At the end the output is generated by GreedyMaxDecoding. That means for each section the most probable token is used for output. By multiplying these probabilities a confidence score for the prediction is obtained.

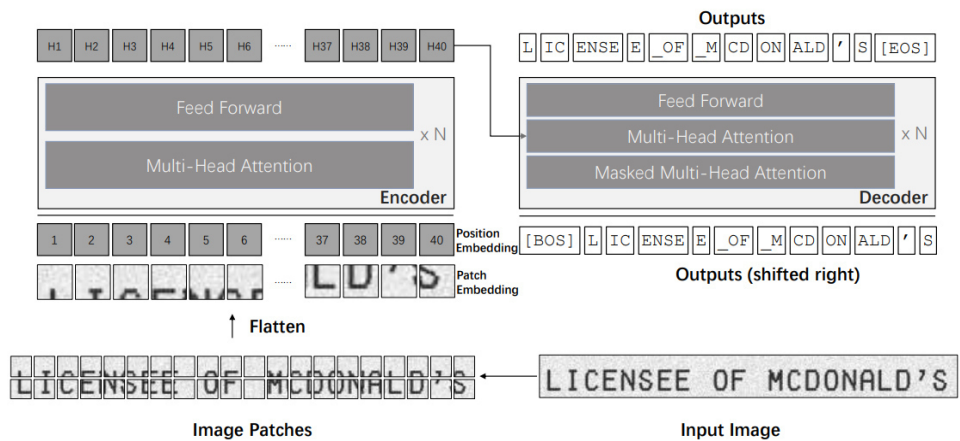

**Figure 4.1:** Model architecture of TrOCR [38]

### 4.2.2 AttentionHTR

AttentionHTR is an attention-based sequence-to-sequence model for handwritten word recognition published by the Uppsala University of Sweden in 2022[39]. The architecture of AttentionHTR consists of four stages: a transformation stage, a feature extraction stage, a sequence modeling stage, and a prediction stage. In the transformation stage, the input word image is normalized via a thin-plate-spline (TPS) transformation, scaling it to a size of 100×32 pixels. The normalized, resized image is then fed into the encoder. The encoder includes the feature extraction stage and the sequence modeling stage. The feature extraction stage consists of a ResNet that encodes the input into a visual feature map. In the sequence modeling stage, this visual feature map is then transformed into a sequence of features and is used by a BLSTM to capture contextual information in the sequence. The output sequence is then used in the prediction phase by an attention-based decoder, which consists of an unidirectional LSTM and content-based attention mechanisms. The decoder outputs a probability matrix with a likelihood for each entry of the character set per sequence step. Then, for each sequence step, the character with the highest probability is used as the prediction (GreedyMaxDecoding). Again, a confidence score can be calculated by multiplying the individual output probabilities.

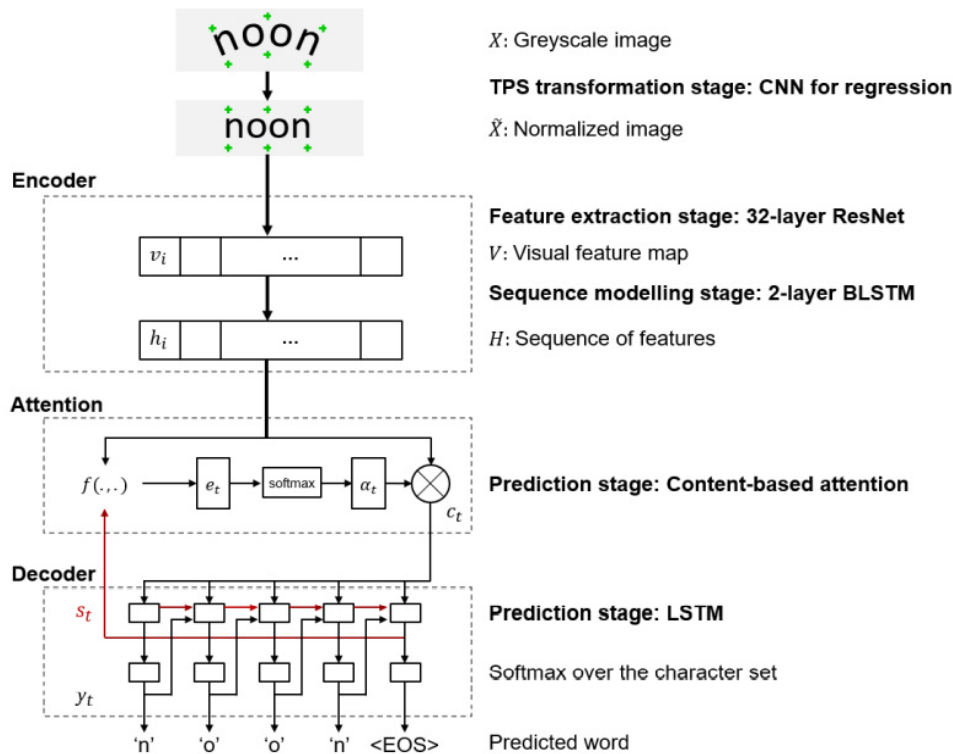

**Figure 4.2:** Model architecture of AttentionHTR [39]

### 4.2.3 SimpleHTR

The last model to be looked at is SimpleHTR, which was published by Harald Scheidl in 2018[40]. The structure of SimpleHTR corresponds to a typical CRNN architecture. This means that the architecture consists of 3 components. The first component is a convolutional neural network (CNN), which receives as input the original image previously reduced to 128×32 pixels and creates a feature sequence matrix from it. The feature sequence matrix is then passed to a recurrent neural network (RNN) component, here given as a LSTM. This component creates a probability matrix of all available characters per image section. Finally, the Connectionist Temporal Classification (CTC) component is used for decoding, again using Greedy-MaxDecoding. A confidence score is also calculated from the probability matrix of the RNN output.

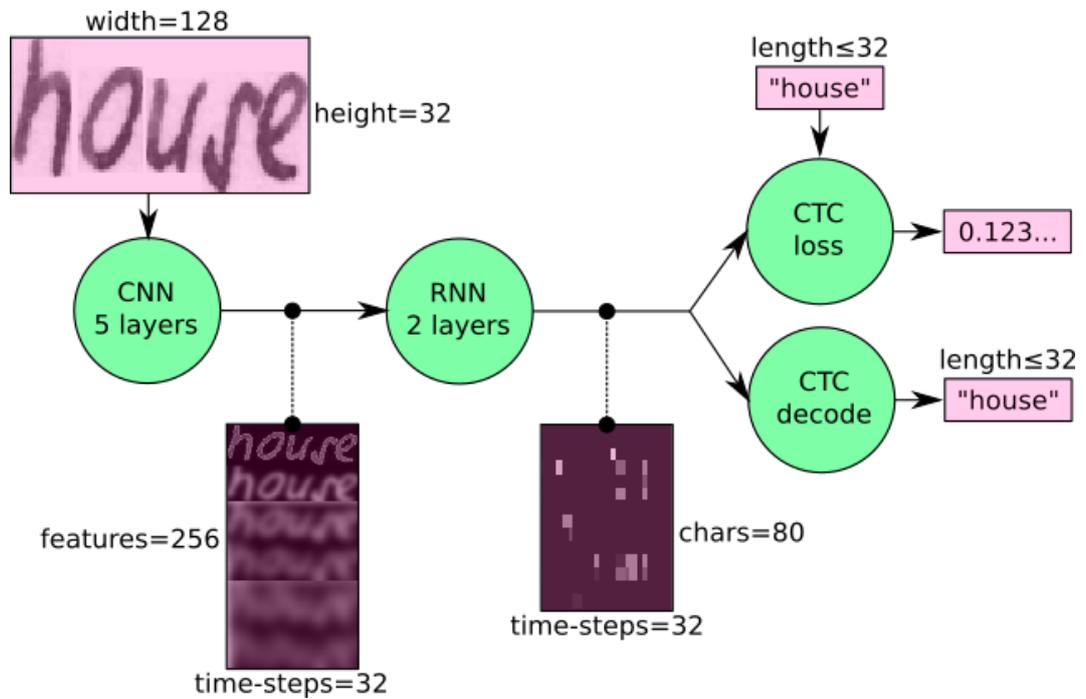

**Figure 4.3:** Model architecture of SimpleHTR [40]

## 4.3  Ensemble Learning Methods

Now the ensemble learning methods considered for OCR shall be mentioned and explained. As described in the background, the construction of an ensemble takes place on 3 levels. At the same time, it should be clarified on which design level which ensemble method can be implemented. Afterwards, other popular ensemble learning methods will be discussed and it will be explained why they are not covered in this work.

### 4.3.1 Dataset Level

At the dataset level, decisions are made about which data is given to which base learner for the training. OCR of course strongly limits the possible ensemble methods. Therefore, only methods that OCR allows as a sequence labeling/classification task are possible. This means that all methods can only be applied on the dataset level if they do not damage the image structure too much, which prevents any feature sampling for example. Nevertheless, there are infinite other ways to split the training data among the base learners. But the most popular and at the same time generic ones are, according to the author's opinion, CompleteData, Bagging, KFOLD and Partitioning.

**CompleteData**   The first and simplest option is to simply give each base learner the complete training data set unchanged. The principle of divesity is then implemented at the base learner level. CompleteData is not a special ensembling method, instead it is the default way if none is implemented at the dataset level.

**Bagging**   One of the best known ensemble learning methods is the bootstrap aggregation method (bagging)[9, 10, 13, 19]. It corresponds to the combination of bootstrapping and a later chosen form of aggregation. In bootstrapping, the original dataset is randomly resampled with replacement for each base learner, usually keeping the length of the original dataset. It is expected that this will diversify the distribution of the training input and counteract a possible poor selection of the training dataset. The possible forms of aggregation are discussed in the output level chapter.

**KFOLD**   In machine learning, k-fold-cross-validation is very popular mainly for the evaluation of machine learning algorithms[41]. In this method, the first step is to merge the train and validation parts within a train/val/test split of the dataset. This now newly combined dataset is then split into k new training and validation splits, where always 1/kth of the dataset becomes the new validation part, which is disjoint to the other k-1 . The remainder then becomes the training dataset. For evaluation, the machine learning algorithm is then retrained on each of the k train/val divisions and tested on it using the unmodified test dataset. The k test results of the individually trained models are then used to evaluate the algorithm. This procedure of the dataset split and training can also be taken advantage of in Ensemble Learning [12, 13]. That way, each baser learner receives one of the k train/val datasets for training. Then, instead of just testing the baser learners, they are applied for prediction and merged in the output layer. Therefore, the number of baser learners goes hand in hand with the number of divisions. For the rest of the work, this approach is referred to with KFOLD.

**Partitioning**    In partitioning, the training dataset is divided among k base learners [9, 10]. This means that each base learner receives a kth disjoint part of the training dataset for training. Especially with large datasets, it is hoped that this will result in a faster training process with consistent results.

### 4.3.2  Base Learner Level

At the base learner level, all decisions are made for the base learners underlying the ensemble. Here, the options available are limited by the base learner architectures chosen. According to the author, the following 3 decisions have to be made: which base learners will be selected, how many base learners will be selected and how to initialize the base learners. These are partly not explicit ensemble learning methods found in the literature, they rather describe the general procedure for the construction of the ensemble on base learner level.

**Base Learner Selection**    As described above, it must first be selected which type of base learner is used. That means, which machine learning algorithm underlies which base learner, here either TrOCR, AttentionHTR or SimpleHTR. There are two approaches for this: the heterogeneous or the homogeneous approach [9] In the homogeneous case, all base learners are based on the same architecture, for example k times TrOCR for k base learners. In the heterogeneous approach, different architectures are used for the base learners, such as k/3 times TrOCR, k/3 times AttentionHTR and k/3 times SimpleHTR for k base learners or in other variations.

**Number of Base Learners**    The next step is one of the most important. It must be chosen how many k base learner will be used at all. This decision is fundamental for the ensemble. If k is chosen too small, the ensemble may have too little diversity of opinions. If k is chosen too large, the computational effort may be too high to be useful in practice. Also, k should be an odd number so that a total majority is more likely in voting procedures.

**Base Learner Initialisation**    Finally, the choice has to be made, which base learner will be initialized with which parameters and which initial weights it will receive. As it is usual also without ensemble learning, the initialization of the initial weights can be done randomly or by transfer learning. Even without different parameter choice/random initialization of the base learner, it is possible that the base learners become diverse enough, because their training does not necessarily have to be deterministic.

### 4.3.3 Output Level

At the output level, the predictions of the base learner are combined for the final result. Since labels/classifaction are combined here, only voting, probability and last layer combination methods are possible. Therefore, all combination methods that use pure numerical predictions as an input, such as averaging, which can be used for regression problems, are not possible. In this work, majority voting, weighted voting and max probability methods are discussed.

**WordVote**   The most popular and at the same time simplest method for merging the predictions is majority voting [9, 13] here called WordVoting. In WordVoting, each of the base learner's predictions is counted by the number of times the prediction occurred. The prediction that occurred most often is then used as the final result. If there is no absolute majority, for example, if all base learners vote differently, the first found maximum is then chosen as the output. This is also the case for the other methods.

**CharVote**   CharVoting follows a similar approach as WordVoting. The main difference is that instead of voting for the whole prediction, each character is voted individually. The idea for this came from the author to see if voting on character level can be more effective. In the implementation, all predictions are first brought to the same length by appending spaces. Now for each position i it is checked which letter was predicted the most. This letter is then at position i of the output. Therefore a voting takes place for each position of the character level. Finally, the excess blanks are truncated.

**WeightedWordVote**   In WeightedWordVoting, as the name implies, the predictions of the k base learners are weighted for voting [19, 20, 24]. The weighting process works as follows. Prior to the predictions, a ranking of the base learners is determined based on their performance on the validation dataset. The predictions of the best performing base learner are then weighted with k points, those of the second best performing base learner with k-1 and so on until the prediction of the worst base learner is weighted with only 1. The weights of all equal predicitons are then added and the prediction with the highest weight is returned as the output.

**WeightedCharVote**   The process of WeightedCharVoting is very similar to that of WeightedWordVoting. But here again the final output is decided on character level. Also in this case it is interesting to see if the voting on total or character level makes a difference. First the predicitons of the base learner are brought to the same length like in the Charvoting with the appending of blanks. Then, for each position of the character level, the weighted voting just described takes place. At the end, any exceeding blanks are shortened.

**MaxProb**   The MaximumProbability method (MaxProb) works with the confidence scores of the base learners for the predictions [19]. Here, the base learner prediction with the highest confidence score is simply returned at the end.

**AvgProb**   In the MaxProb method, identical predictions are only considered individually. The AveragedMaximumProbability method (AvgProb) enhances the MaxProb method. Here the goal was to see if the combination of the confidence scores of the same predictions makes a difference. Again, the idea came from the author of this paper. From all identical base learner predictions the average of the confidence scores is calculated. Just like in the MaxProb procedure, the prediction with the highest confidence score is then returned.

### 4.3.4  Combining the Methods

Finally, the question obviously arises: How exactly are the above-mentioned ensemble methods of the design levels combined? The answer is quite simple. Every method mentioned above can be combined with all methods of the other design levels. The implementation of this will be considered in more detail in the experiments.

### 4.3.5  Ineligible Methods

There are, of course, other methods that are theoretically possible in combination with OCR. However, these have not been dealt with in this work because they are too application-specific or cannot be combined with the selected OCR models. However, for further research purposes, other very popular methods or possibilities that are interesting from the author's point of view will be mentioned here.

On the data set level, Input Variation and Category Sort methods are worth mentioning. Input Variation varies the input for the base learner to achieve more diversity [9]. For example, the base learners receive differently blurred image data. In category sort methods the training data is sorted by certain categories and each base learner receives certain categories for the training, for example only numbers. The idea for this came up while creating this work, since the medical data consists of different categories. However, both methods are very dataset specific and therefore difficult to evaluate in general. In addition, Input Variation is very similar to DataAugmentation, which is a big topic for its own.

On base learner level the ensemble methods are limited by the selected OCR models. So any possibilities are excluded, that can not be combined with the 3 previously mentioned OCR models. Worth mentioning here are the very popular Boosting method or Fast Ensemble methods like Snapshoting. Boosting uses weighted learning algorithms to more strongly influence the incorrectly predicted training data of the previous base learner when training the next base learner [9]. Fast Ensemble methods aim to compensate for the training disadvantages of ensembles with specially designed learning algorithms[24]. The most famous example Snapshoting creates an ensemble from all minima it finds in the training process, for which cyclic learning rates are needed[24]. However, the 3 models considered in this paper do not possess the necessary properties for the implementation of the above mentioned examples. It is also very common to let base learners of the same architecture vary in their parameters e.g. by different number of layers [13]. However, here the decision was made not to do this, since on the one hand this is again very application-specific and on the other hand the developers of the models already concerned themselves enough with the selection for the correct parameters.

On the output level one of the most important ensemble methods is stacking. Stacking involves the training of an extra machine learning model that combines the predictions or last layers of the base learner [9, 10, 13]. Unfortunately, stacking is not very generic, as it has to be adapted very precisely to the model structures. Moreover, it is such an extensive topic that an extra paper would be necessary. In theory, the Bayes Combination Method for the output layer appears repeatedly [12]. It is considered to be the most accurate method in theory, but it is not practically feasible here, since knowledge about the probabilities of occurrence of the labels would be needed. Other very popular and interesting methods are Classifier Selection Methods. Here only certain "experts" of the base learners are used for the ensemble prediction, e.g. experts only for numbers[24]. This can be done using the Category Sort methods or by algorithms that evaluate the base learners for different problems. The Classifier Selection Methods were also not covered, because they are again a very large topic and very application specific.

# 5 Experiments

Next, the practical procedure will be explained. First of all, it will be clarified on which datasets the experiments take place and which measured values are recorded. Then the experimental setup/execution will be described and at the end the final results will be presented.

## 5.1 Datasets

For the experiments we will use 2 datasets with handwritten images.

**Duke Dataset** The first is the Duke Dataset already mentioned in the introduction, which consists of cell entries from medical tables. The data were collected on special patient records in the Rice Diet study by Walter Kempner for the treatment of hypertension [42]. The image data now available were thereby extracted from 5 columns of these records. These were the blood pressure column, weight column, and sodium, chlorine, potassium columns of the patients urine. The total size of the dataset is 6287. There are a total of 2367 cell images of the blood pressure column, 2083 of the weight column, 151 of the sodium urine column, 764 of the chloride urine column, and 79 of the potassium urine column. In addition, the dataset also consists of 843 empty cell images. The following characters occur in the labels of the images: ",./0123456789 $\frac{1}{4} \frac{1}{2} \frac{3}{4} \frac{1}{8}$"

**IAM Words Dataset** The second dataset is the very popular IAM Words Dataset from the IAM Handwriting Database[43]. In the OCR field, the IAM Words Dataset is regularly used to evaluate new technologies. It was published by the University of Bern and contains 115320 isolated and labeled english word images. These word images were extracted with automatic mechansims from scans of english scripts, were manually verified and are now available in gray tones. The following characters appear in the labels of the images:" 0123456789abcdefghijklmnopqrstuvwxyzABCDEFGHIJKLMNOPQRSTUVWXYZ!" &.'()*+,-./:;?". In the following, the IAM Word Dataset will also just be called the IAM Dataset.

**Split Ratio** Both datasets were divided into a train, validation and test part for the experiments. The reason for this is that the OCR methods require a disjoint set of training and validation data for the learning process. On the other hand, at the end, the methods need to be evaluated with the test data that has never been observed before. Since the Duke Dataset is relatively small, a large test dataset is necessary to evaluate the ensemble learning techniques well. For a good result of the ensembles the training and validation part must not be too small. Therefore, the decision was made for a 70/15/15 split ratio, because it best combines the above requirements. For the Duke Dataset, the number of training images is therefore 4400, for the validation part 943 and for the test part 944. For the IAM Dataset, there are 80722 images for training, 17298 for validation, and 17298 for testing. Hence, the IAM Dataset is about 18 times larger than the Duke Dataset.

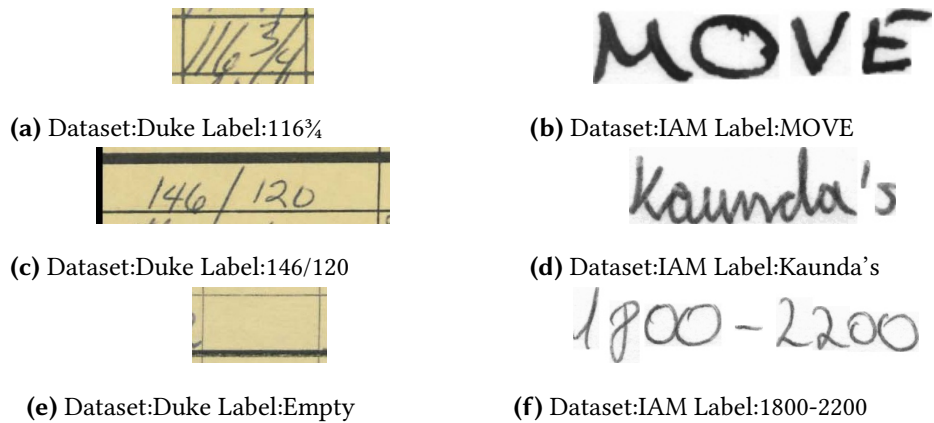

**(a)** Dataset:Duke Label:116¾  **(b)** Dataset:IAM Label:MOVE

**(c)** Dataset:Duke Label:146/120  **(d)** Dataset:IAM Label:Kaunda's

**(e)** Dataset:Duke Label:Empty  **(f)** Dataset:IAM Label:1800-2200

**Figure 5.1:** Example pictures from the 2 Datasets with Duke on the left and IAM on the right side

## 5.2 Metrics

To measure the accuracy of the predictions on the datasets, the Word Accuracy (WA) and the Char Error Rate (CER) are used. These metrics are commonly used to measure OCR methods, including the OCR models discussed here [38, 39, 40]. Word Accuracy is the percentage of correctly estimated words out of the total number of input words. It is calculated by dividing the number of correctly estimated words (len(CorrectWords)) by the number of all input words (len(Inputs)) multiplied by 100. The Char Error Rate (CER) corresponds to the normalized levenstein distance of a word multiplied by 100[38]. The levenstein distance is the number of insertions, substitutions, and deletions of characters to obtain the true word. This is normalized by dividing throug number of characters of the ground truth. For measurement the average of all CER's of the predictions will be used. However, for the evaluation of the experiments only the Word Accuracy is used, since this was sufficient for the evaluation of the research questions. For those interested, the CER's will be part of the appendix.

At this point it should be mentioned that the diversity principle explained in the background chapter is not really measurable. Although some diversity metrics have been designed, there is no agreement on a standard. In fact, it is generally agreed that there is no correlation between diversity metrics and the prediction performance of ensembles [44]. This does not mean that the principle of diversity is invalid. The base learners should still be designed to be diverse. However, diversity is difficult to measure since methods can also be diversely designed and lead to the same result.

$$WordAccuracy = len(CorrectWords)/len(Inputs) * 100$$

**Figure 5.2:** Word Accuracy

## 5.3 Experimental Setup and Execution

It shall now be clarified how to proceed exactly in order to evaluate the ensemble methods mentioned in the methodology.

**Initialisation of OCR Modells**   Before each experiment, the OCR models must be initialized first. This is done as follows.

**TrOCR**   For TrOCR the pre-trained "trocrbasestage1" model is used. For handwriting recognition the "trocrbasehandwritten1" model would be more recommended. However, this one has already been pre-trained on the IAM Dataset. So the decision was made to use the TrOCR default model from the TrOCR paper. Nevertheless "trocrbasehandwritten1" is mentioned here, because it will give better results for other OCR projects. Otherwise, the configuration of the model is largely untouched. The only parameters that are still set individually are number of workers with value 8, batch size with value 20 and train epochs with value 8, which are important for the training process. The number of train epochs is not very high, because TrOCR does not need many epochs for a good finetuning. The choice for batch size and number of workers was made because this worked best with the available resources. However, the setting of the parameter values for the models has no strong relevance for this work. They only have to be constant for all experiments to provide a comparable basis for the ensemble methods. Additionally it should be mentioned that TrOCR does not need a character list, because the standard token dictionary is sufficient here.

**AttentionHTR**   AttentionHTR is initialized with the pre-trained "AttentionHTR Imgur5K sensitive" model. It is used because on the one hand case sensitive labels are in the data and on the other hand only this model provided by the developers is not already pre-trained on IAM. Additionally, the values for batch size and character list will be set. Otherwise the parameters set by AttentionHTR will be left as they are. Again, the relevance of these parameters is relatively insignificant, since the main focus lies on the ensemble methods. For the batch size the value 32 was chosen and for the character list the above mentioned character list of the respective dataset is used. The choice for the batch size was also made here due to the best usage of the available resources.

**SimpleHTR**   SimpleHTR is initialized without a pre-trained model, since there is no model with the chararcter list required for the Duke Dataset. The character list influences the parameters of the architecture so strongly that the use of the model is no longer possible. In order to be able to draw better comparisons later, no pre-trained model is also used during the experiments with the IAM Dataset. In addition, the values for batch size and earlystopping are set to 100. Independent experiments showed that these provided the best values.

**Experimental Execution**   Now the execution of the experiments can be considered. First of all, the 3 OCR models will be evaluated separately on the Duke and IAM Datasets. For this purpose, the training of the initialized models is performed with the help of the train/val dataset. Once the training is complete, the Word Accuracy/Char Error Rate of the models is calculated on the test dataset. This is already the experiment for the evaluation of the single models. The experimental procedure for the ensemble methods is as follows. Each experiment, which was performed on the Duke Dataset, should also be performed on the IAM Dataset. For this purpose, the datasets are first prepared using the 4 dataset level methods described in the methodology. The training of the base learners is now to be performed with the datasets that have been processed. For this, the last fundamental decisions have to be made on the base learner level, i.e. which base learner corresponds to which OCR model and how many base learners make up the ensemble. For the base learner selection, both the homogeneous approach with all OCR models and the heterogeneous approach with an equally distributed number of OCR models will be considered. For the evaluation of the homogeneous ensembles, k=5 is chosen, because k=3 would be too few predictions and k=7 would require too much computational effort. Also an odd number is important. For the evaluation of the heterogeneous ensemble k=9 is chosen with 3 models each of TrOCR, AttentionHTR and SimpleHTR. In each case the first 3 trained base learners of the homogeneous ensembles will be selected here. This means that the dataset level methods KFOLD and Partitioning are of course not implemented correctly at this point. But this is not important here, because it is more about the question whether the heterogeneous approach brings more added value compared to the homogeneous approach. The choice for k=9 was made because the ratio of the different OCR models should be the same. In addition, k=3 is too small, k=6 even and k=12 too large. Now the training of the base learners can be triggered using the previously processed train/val datasets. Once the training is completed, the Word Accuracy/Char Error Rate rate of the base learners is calculated on the test dataset. Finally, the base learners are combined using the output level methods and their Word Accuracy/Char Error Rate is also calculated. In total, there are 32 different ensembles for the evaluation of the methods.

## 5.4 Results

From the above described experiments, the following Word Accuracy values were obtained. The values for the average Char Error Rate can be found in the appendix for those who are interested. The number of results is overwhelming at first. But in the conclusions chapter these values are broken down to more simple levels. In the tables, the best measured value is always printed in bold, as well as for ensembles best base learner.

|  | TrOCR | AttentionHTR | SimpleHTR |
|---|---|---|---|
| Duke | **96,82** | 93,43 | 84 |
| IAM | 72,83 | **86,84** | 75,74 |

**Table 5.1:** Experimentation Results of Single OCR Modells

| Dataset Methods | Base Learners | | | | | Output Level Method | | | | | |
|---|---|---|---|---|---|---|---|---|---|---|---|
| | 1 | 2 | 3 | 4 | 5 | WordVote | CharVote | Weighted-WordVote | Weighted-CharVote | MaxProb | AvgProb |
| Complete Data | 90,67 | **96,29** | 88,98 | 86,02 | 96,08 | 95,86 | 95,44 | **97,03** | 93,33 | 91,38 | 83,41 |
| Bagging | 93,54 | 92,69 | 95,34 | **95,66** | 95,02 | 96,61 | 95,44 | **96,93** | 95,02 | 93,31 | 92,19 |
| K-FoldCross | 96,34 | 96,4 | 96,73 | 88,87 | **96,82** | 98,09 | 96,82 | 97,88 | 96,50 | 93,03 | 88,22 |
| Partitioning | 0,1 | 0,2 | 0,0 | 0,0 | 0,32 | 0,10 | 0,0, | 0,11 | 0,0 | 0,32 | 0,32 |

**Table 5.2:** Experimentation Results of homogenous TrOCR Ensembles on Duke Data

| Dataset Methods | Base Learners | | | | | Output Level Method | | | | | |
|---|---|---|---|---|---|---|---|---|---|---|---|
| | 1 | 2 | 3 | 4 | 5 | WordVote | CharVote | Weighted-WordVote | Weighted-CharVote | MaxProb | AvgProb |
| Complete Data | **87,43** | 77,62 | 75,24 | 76,24 | 80,45 | 83,51 | 76,08 | 83,51 | 74,57 | **85,5** | 82,56 |
| Bagging | 78,98 | **79,08** | 68,48 | 70,69 | 72,16 | **80,29** | 67,18 | **80,29** | 65,99 | 78,80 | 74,69 |
| K-FoldCross | 78,89 | 76,28 | **81,11** | 70,74 | 72,52 | **80,57** | 70,63 | 80,5 | 69,8 | 79,32 | 76,64 |
| Partitioning | 74,13 | **79,25** | 49,87 | 77,98 | 76,43 | **81,17** | 64,73 | 81,17 | 63,13 | 79,44 | 75,54 |

**Table 5.3:** Experimentation Results of homogenous TrOCR Ensembles on IAM Data

| Dataset Methods | Base Learners | | | | | Output Level Method | | | | | |
|---|---|---|---|---|---|---|---|---|---|---|---|
| | 1 | 2 | 3 | 4 | 5 | WordVote | CharVote | Weighted-WordVote | Weighted-CharVote | MaxProb | AvgProb |
| Complete Data | **94,07** | 92,27 | 93,96 | 93,75 | 92,26 | 95,76 | 94,28 | 95,76 | 94,28 | **96,08** | 94,60 |
| Bagging | 89,62 | 90,78 | 90,89 | **91,53** | 89,94 | **93,96** | 91,42 | 93,75 | 91,74 | 93,11 | 91,84 |
| K-FoldCross | 92,58 | **93,11** | 91,00 | 92,48 | 91,84 | **96,40** | 94,28 | 95,87 | 93,75 | 95,97 | 93,54 |
| Partitioning | 60,38 | 65,78 | 69,39 | 63,35 | **69,49** | 75,42 | 71,08 | **76,48** | 67,69 | 76,38 | 69,60 |

**Table 5.4:** Experimentation Results of homogenous AttentionHTR Ensembles on Duke Data

| Dataset Methods | Base Learners | | | | | Output Level Method | | | | | |
|---|---|---|---|---|---|---|---|---|---|---|---|
| | 1 | 2 | 3 | 4 | 5 | WordVote | CharVote | Weighted-WordVote | Weighted-CharVote | MaxProb | AvgProb |
| Complete Data | 87,03 | 87,43 | **87,63** | 87,07 | 86,47 | 89,45 | 87,37 | 89,32 | 87,18 | **89,80** | 89,09 |
| Bagging | 85,33 | 85,79 | **86,69** | 86,39 | 85,63 | 88,68 | 86,23 | 88,46 | 85,83 | **88,98** | 88,29 |
| K-FoldCross | 86,96 | 86,68 | 86,96 | **87,06** | 86,83 | 89,40 | 87,42 | 88,99 | 87,13 | **89,84** | 89,03 |
| Partitioning | **84,02** | 83,63 | 83,19 | 83,77 | 83,90 | 86,83 | 84,14 | 86,57 | 83,72 | **87,19** | 86,23 |

**Table 5.5:** Experimentation Results of homogenous AttentionHTR Ensembles on IAM Data

| Dataset Methods | Base Learners | | | | | Output Level Method | | | | | |
|---|---|---|---|---|---|---|---|---|---|---|---|
| | 1 | 2 | 3 | 4 | 5 | WordVote | CharVote | Weighted-WordVote | Weighted-CharVote | MaxProb | AvgProb |
| Complete Data | 84,85 | 82,63 | 83,69 | **87,29** | 82,42 | **91,10** | 84,96 | 90,36 | 89,41 | 90,36 | 86,44 |
| Bagging | 85,06 | 78,18 | **86,97** | 83,90 | 84,53 | 89,41 | 84,85 | 89,51 | 87,92 | **90,68** | 88,14 |
| K-FoldCross | **91,21** | 89,09 | 90,57 | 87,71 | 91,00 | 94,49 | 90,78 | 93,22 | 90,15 | **94,92** | 91,84 |
| Partitioning | 55,40 | 43,40 | **58,26** | 50,42 | 45,33 | **64,60** | 59,53 | 63,98 | 57,94 | 61,44 | 55,93 |

**Table 5.6:** Experimentation Results of homogenous SimpleHTR Ensembles on Duke Data

| Dataset Methods | Base Learners | | | | | Output Level Method | | | | | |
|---|---|---|---|---|---|---|---|---|---|---|---|
| | 1 | 2 | 3 | 4 | 5 | WordVote | CharVote | Weighted-WordVote | Weighted-CharVote | MaxProb | AvgProb |
| Complete Data | **74,66** | 74,25 | 74,01 | 73,46 | 73,65 | 80,44 | 72,45 | 79,25 | 71,76 | **80,73** | 77,28 |
| Bagging | 74,38 | **75,04** | 73,24 | 73,72 | 73,18 | 80,47 | 73,01 | 79,58 | 72,35 | **80,61** | 77,53 |
| K-FoldCross | 76,22 | **76,73** | 76,43 | 76,22 | 76,52 | 82,28 | 75,76 | 81,27 | 75,56 | **82,40** | 79,69 |
| Partitioning | 63,91 | 62,60 | 63,79 | **64,74** | 64,57 | 71,98 | 62,87 | 71,44 | 62,15 | **73,04** | 68,48 |

**Table 5.7:** Experimentation Results of homogenous SimpleHTR Ensembles on IAM Data

| Dataset Methods | Output Level Method | | | | | |
|---|---|---|---|---|---|---|
| | WordVote | CharVote | Weighted-WordVote | Weighted-CharVote | MaxProb | AvgProb |
| Complete Data | 97,35 | 92,48 | **98,09** | 91,95 | 92,01 | 81,35 |
| Bagging | 96,50 | 88,87 | **97,03** | 87,61 | 93,06 | 87,98 |
| K-FoldCross | **97,99** | 93,75 | 96,61 | 93,75 | 97,35 | 91,95 |
| Partitioning | 75,10 | 36,33 | **76,48** | 7,20 | 33,26 | 23,09 |

**Table 5.8:** Experimentation Results of heterogenous Ensembles on Duke Data

| Dataset Methods | Output Level Method | | | | | |
|---|---|---|---|---|---|---|
| | WordVote | CharVote | Weighted-WordVote | Weighted-CharVote | MaxProb | AvgProb |
| Complete Data | **90,70** | 76,42 | 89,70 | 74,74 | 88,67 | 80,52 |
| Bagging | **89,59** | 72,39 | 88,91 | 69,54 | 85,92 | 80,84 |
| K-FoldCross | **90,12** | 77,51 | 89,95 | 74,47 | 87,84 | 81,16 |
| Partitioning | **87,61** | 61,62 | 86,72 | 59,67 | 83,19 | 71,90 |

**Table 5.9:** Experimentation Results of heterogenous Ensembles on IAM Data

# 6 Discussion

With the measured values collected above, the research questions will now be answered.

## 6.1 Research Question 1

First of all, concerning the answer of RQ1: "Can Ensemble Learning improve the Accuracy of modern OCR methods?". The idea to clarify this question was to compare the models trained alone with the ensemble results. In order to not confront 24 measured values with only one single one, only the Word Accuracies of the highest Word Accuracies of the respective ensemble are considered in the following table 6.1. Thereby it will be looked whether the Word Accuracies of the highest ensemble method is also larger than that of the single model. The values of the heterogeneous ensembles are not considered here, since these consist of all OCR models and therefore can hardly be compared with the single model values.

It can be seen immediately that the Word Accuracy of the ensembles is always better than the ones of the single models. Of course, the relation is quite unbalanced here. The reason is that each single WA is evaluated alone with the maximum of 24 output level combinations. In order to balance this a bit, the base learners should also be included, because the base learners also obviously correspond to a single model. Therefore the highest Word Accuracy value of the respective base learner or the single model should be used for the comparison. Hence, in Table 6.2 per OCR model/dataset combination, 21 single models are now compared with the 24 output level combinations.

Even with the inclusion of the base learners, the individual models were only once better than one of the ensembles. For the ensemble that is worse than the base learner (TrocrIamKFOLD1), it is noticeable that the principle of the highest possible accuracy is violated here. Even with the inclusion of the base learners, the individual models were only once better than one of the ensembles. For the ensemble that is worse than the base learner (TrocrIamKFOLD1), it is noticeable that the principle of the highest possible accuracy is violated here. Since the other base learners of the TrocrIamKFOLD ensemble are substantially more inefficient on the test dataset than the TrocrIamKFOLD1 base learner. If this would be remedied by retraining the other base learners and keeping only the trained base learners that also have higher accuracy, it is very possible that the ensemble would predict better. Furthermore, just because a base learner performs better on the test dataset it does not mean that the ensemble as a whole is worse. It is not known to what extent the base learner generalizes well. It is quite possible that this base learner performs well only here on the test dataset and is much worse on other unknown data. As mentioned in the background, ensembles have the advantage to generalize better and to have a lower risk of overfitting. For this reason, the ensemble would still be preferable here.

With this knowledge, RQ1 can be answered as follows. Yes, Ensemble Learning can help with high probability to improve the accuracy of OCR methods. However, as seen above, this does not mean that this is always the case. With a good choice of base learners, however, Ensemble Learning is definitely suitable as a possible tool to get a few last percentage points of Word Accuracies. Especially with the background thought of the possible better generalization of the ensembles.

| OCR Modell | Dataset | Single WA | Best Ensemble WA | Single < Best Ensemble ? |
| --- | --- | --- | --- | --- |
| TrOCR | Duke | 96,82 | 98,09 | True |
| AttentionHTR | Duke | 93,43 | 96,40 | True |
| SimpleHTR | Duke | 84,00 | 94,92 | True |
| TrOCR | IAM | 72,83 | 85,53 | True |
| AttentionHTR | IAM | 86,84 | 89,84 | True |
| SimpleHTR | IAM | 75,74 | 82,40 | True |

**Table 6.1:** Comparison of Single WA's with maximal achieved Ensemble WA's

| OCR Modell | Dataset | Best Single or Base Learner WA | Best Ensemble WA | Single < Best Ensemble ? |
|---|---|---|---|---|
| TrOCR | Duke | 96,82 | 98,09 | True |
| AttentionHTR | Duke | 94,07 | 96,40 | True |
| SimpleHTR | Duke | 91,21 | 94,92 | True |
| TrOCR | IAM | 87,43 | 85,53 | False |
| AttentionHTR | IAM | 87,63 | 89,84 | True |
| SimpleHTR | IAM | 76,73 | 82,40 | True |

**Table 6.2:** Comparison of best achieved base learner or single WA's with maximal achieved Ensemble WA's

## 6.2 Research Question 2

In RQ2, the question was asked, "Which Ensemble Learning methods are the most valuable?" To get an answer to this question, the ensemble methods of the different design levels will be compared to find the most performant methods.

### 6.2.1 Dataset Level

For the dataset methods, the decisions of the other design levels and the choice of the dataset obviously have a large impact on them. Therefore, pure Word Accuracies can not be considered here. Instead, it will be counted which method on which model/dataset combination achieved the best possible Accuracy of any output method. This can be seen in table 6.3 In this way, it should be shown which method has performed best and how often.
According to the table it can be assumed that only CompleteDataset and KFOLD provide an additional value. However, only one of the 6 output level methods was counted here. The values of these lie partly very close together, with which this consideration could easily falsify itself. Therefore, the above counting shall take place again for all output methods, which is to be examined in table 6.3
With this data it can now be clearly said that, considering the previously set values, the CompleteData and KFOLD approaches add the most value. This seems to be the case more often for KFOLD than for the CompleteData method. This confirms itself again with view on the total averages of the individual ensemble methods, which can be found in table 6.5

Here, KFOLD also has the highest average accuracy, followed by the CompleteData method. This shows that KFOLD performed with the best results here. For the other 2 DatasetLevel methods, it is immediately noticeable that Partitioning achieved rather unusable values. On TrOCR even so badly that both base learners and output methods reach an Accuracy of nearly and/or equal 0. It is interesting that Partitioning on IAM TrOCR delivers again acceptable results. Therefore, it can be speculated that the number of training data or the number of training epochs is too small for Duke TrOCR. Hence, Partitioning may be a good solution for even much larger datasets or for a different number of base learners. Under the circumstances considered above, however, Partitioning is rather to be discarded. Looking at the average word accuracy of the Bagging method, it is noticeable that it performs only 2 to 1 percentage points worse than the KFOLD or CompleteData methods. Also in the table 6.4 Bagging were able to achieve 2 times the best result of an output level method within a OCR model/dataset combination. This shows that Bagging does not work badly here perse. Nevertheless, KFOLD and CompleData performed much more often and overall better. This could be due to the fact that the bagging intensity is possibly too low, too high or bagging simply works better with more base learners.Still, the potential of the bagging method is proven here. Finally, it can be said for the methods of the dataset level that all of them except for Partitioning delivered good results, whereby the KFOLD and the CompleteData method were the most useful.

| OCR Modell | Dataset | CompleteData | KFOLD | Bagging | Partitioning |
|---|---|---|---|---|---|
| TrOCR | Duke |  | 1 |  |  |
| AttentionHTR | Duke |  | 1 |  |  |
| SimpleHTR | Duke |  | 1 |  |  |
| heterogenous | Duke | 1 |  |  |  |
| TrOCR | IAM | 1 |  |  |  |
| AttentionHTR | IAM |  | 1 |  |  |
| SimpleHTR | IAM |  | 1 |  |  |
| heterogenous | IAM | 1 |  |  |  |
| Complete Count |  | 3 | 5 |  |  |

**Table 6.3:** counted number of best performing dataset level methods per OCR model/dataset combination

| OCR Modell | Dataset | CompleteData | KFOLD | Bagging | Partitioning |
|---|---|---|---|---|---|
| TrOCR | Duke | | 4 | 2 | |
| AttentionHTR | Duke | 4 | 3 | | |
| SimpleHTR | Duke | | 6 | | |
| heterogenous | Duke | 1 | 5 | | |
| TrOCR | IAM | 6 | | | |
| AttentionHTR | IAM | 4 | 2 | | |
| SimpleHTR | IAM | | 6 | | |
| heterogenous | IAM | 3 | 3 | | |
| Complete Count | | 18 | 29 | 2 | |

**Table 6.4:** counted number of best performing dataset level methods per OCR model/dataset combination per output level method

| CompleteData | KFOLD | Bagging | Partitioning |
|---|---|---|---|
| 87,37 | 88,22 | 86,07 | 61,12 |

**Table 6.5:** Averages of the dataset level methods

### 6.2.2 Base Learner Level

On the base learner level, only the heterogenous and homogenous approach as well as the OCR models underlying the base learners can be evaluated, meaning which OCR model is best suited as a base learner. The remaining decisions were made for the experiments as a whole and are not comparable here, because there is too little measurement data or the underlying methodologies are too different, like for example the number of base learners. It is meaningless to compare the number of base learners of 5 with 9 if the methodology is different (homogenous vs heterogenous) and only 2 different numbers are available.

**Comparison of Heterogenous and Homogenous**   For the homogenous vs heterogenous comparison, the maximum Word Accuracies of all homogenous ensembles should first be compared to the maximum Word Accuracies of all heterogenous ensembles on both the Duke and IAM datasets. This can be seen in the following table 6.6.

It is noticeable that there is hardly a difference in the maximum of both approaches. In order to look at a few more values, the highest Word Accuracies of the respective dataset methods and output methods should now also be used, see Table 6.7 and 6.8. The difference between the heterogeneous and the homogenous shall be looked at here. The reason why in the table only the methods WordVote, WeightedWordVote and MaxProb are shown, follows in the Output Level section.

When looking at the differences in the two tables, it is immediately evident that none of the methods stands out. The differences between the two are only a few percentage points apart. The average difference is also just 0.38%. This means that the heterogeneous approach is slightly better here. But this number is so small that this should be given little importance. Thus it can be said that both the homogenous and the heterogenous approach can deliver good results. Nevertheless, it cannot be said here that one of the methods is better. Beside that, it is interesting to mention here that the heterogenous Duke Partitioning Ensemble managed to compensate the weak TrOCR Partitioning base learners. This confirms the advantage of ensembles that bad base learners can be outvoted.

| Dataset | homogenous | heterogenous |
|---------|------------|--------------|
| Duke    | 98.09      | 98.09        |
| IAM     | 89.843     | 90.698       |

**Table 6.6:** Maximum values of homogenous and heterogenous Ensembles

| Dataset | Dataset Level Methods | Maximum WA's homogenous | Maximum WA's heterogenous | Difference |
|---|---|---|---|---|
| Duke | CompleteData | 97,03 | 98,09 | 1,06 |
| Duke | Bagging | 96,93 | 97,03 | 0,1 |
| Duke | KFOLD | 98,09 | 97,99 | -0,1 |
| Duke | Partitoning | 76,48 | 76,48 | 0 |
| IAM | CompleteData | 89,8 | 90,7 | 0,9 |
| IAM | Bagging | 88,98 | 89,59 | 0,61 |
| IAM | KFOLD | 89,84 | 90,12 | 0,28 |
| IAM | Partitoning | 87,19 | 87,61 | 0,42 |

**Table 6.7:** Differences of the highest achieved dataset level methods of the heterogenous and the homogenous approach

| Dataset | Dataset Level Methods | Maximum WA's homogenous | Maximum WA's heterogenous | Difference |
|---|---|---|---|---|
| Duke | WordVote | 98,09 | 97,99 | -0,10 |
| Duke | WeightedWordVote | 97,88 | 98,09 | 0,21 |
| Duke | MaxProb | 96,08 | 97,35 | 1,27 |
| IAM | WordVote | 89,45 | 90,7 | 1,25 |
| IAM | WeightedWordVote | 89,32 | 89,95 | 0,63 |
| IAM | MaxProb | 89,84 | 88,67 | -1,17 |

**Table 6.8:** Differences of the highest achieved output level methods of the heterogenous and the homogenous approach

**Base Learner Selection**   Now it will be evaluated which OCR model brings the most additional value as a base learner. In table 6.9 the averaged Word Accuracies without the by outlayers affected Partitioning can be seen. In Table 6.10 the maximum achieved Word Accuracies are visible. The values of the heterogenous approach are again not considered here, as well as in RQ1.

The first thing that stands out is that SimpleHTR is much worse than TrOCR and AttentionHTR on average and at maximum. This makes sense because SimpleHTR is the oldest architecture and does not use a pre-trained model for the initialization. On the Duke Dataset, with the above initializations TrOCR and AttentionHTR predicted about the same on average. Nevertheless, here TrOCR has the largest maximum, which is 1.69% higher than that of AttentionHTR, This is different on the IAM Dataset. Here AttentionHTR is much more accurate in the average as well as in the maximum. Thus, in conclusion it can be said that TrOCR is best suited for the Duke Dataset and AttentionHTR is best suited for the IAM Dataset. Of course, this is only in consideration of the decisions made before.

| Dataset  | TrOCR | AttentionHTR | SimpleHTR |
|----------|-------|--------------|-----------|
| Duke     | 94,25 | 94,24        | 89,91     |
| IAM      | 77,25 | 88,36        | 77,91     |
| Combined | 85,75 | 91,3         | 83,92     |

**Table 6.9:** Averaged WA's of base learners without Partitioning

| Dataset | TrOCR | AttentionHTR | SimpleHTR |
|---------|-------|--------------|-----------|
| Duke    | 98.09 | 96.4         | 94.92     |
| IAM     | 85.53 | 89.84        | 82.4      |

**Table 6.10:** Maximum reached WA's of base learners

### 6.2.3  Output Level

Last but not least, a statement should be made about the output methods. The first thing that stands out is that voting at the character level or the AvgProb method do not add any value here. The Word Accuracy of these methods are always worse than their counterpart, the voting on word level or the standard MaxProb method. For this reason, they are not being looked at in the further evaluation. In the case of voting at character level, however, it can be assumed that these methods may be better at dealing with longer words or sentences, because the datasets only have quite short labels.

Like when looking at the Data Set Level methods, it will now be counted which method has achieved the best possible accuracy of any dataset level method on which model/dataset combination. This can be seen in table 6.11

Looking at these results, it can be assumed that mainly WordVoting and MaxProb work better. In order to have more values for the comparison, the highest output level method will be counted again for all dataset level methods, which is to be examined in table 6.12 (excluded the values of Partition Duke TrOCR)

Once again, WordVote and MaxProb were much better than the WeightedWordVoting method. But some peculiarities can be seen here. For SimpleHTR, the MaxProb procedure performed significantly better. In general, for the homogenous ensemble on the IAM Dataset, the MaxProb method always gave the best result see Table 6.11 . For the predictions on the Duke Dataset, as well as for TrOCR and the heterogenous ensembles, the voting methods achieved the best Word Accuracies. Therefore, it can be said that the methods predict better depending on the circumstances. To the WeightedWordVoting procedure it can be noted that this can lead also to the best result. When looking at the average accuracy of the methods, it can be seen that all 3 methods are not very far apart, see Table 6.13 Finally, on the output level, it can be said that all 3 methods can lead to good results. However, WordVoting and MaxProb achieved the best results.

| OCR Modell | Dataset | WordVote | WeightedWordVote | MaxProb |
|---|---|---|---|---|
| TrOCR | Duke | 1 | | |
| AttentionHTR | Duke | 1 | | |
| SimpleHTR | Duke | | | 1 |
| heterogenous | Duke | | 1 | |
| TrOCR | IAM | | | 1 |
| AttentionHTR | IAM | | | 1 |
| SimpleHTR | IAM | | | 1 |
| heterogenous | IAM | 1 | | |
| Complete Count | | 3 | 1 | 4 |

**Table 6.11:** Counted number of best performing output level methods per OCR model/dataset combination

| OCR Modell | Dataset | WordVote | WeightedWordVote | MaxProb |
|---|---|---|---|---|
| TrOCR | Duke | 1 | 2 | |
| AttentionHTR | Duke | 2 | 1 | 1 |
| SimpleHTR | Duke | 2 | | 2 |
| heterogenous | Duke | 1 | 2 | |
| TrOCR | IAM | 3 | 1 | 1 |
| AttentionHTR | IAM | | | 4 |
| SimpleHTR | IAM | | | 4 |
| heterogenous | IAM | 4 | | |
| Complete Count | | 14 | 6 | 11 |

**Table 6.12:** Counted number of best performing output level methods per OCR model/dataset combination per dataset level method

| WordVoter | WeightedWordVote | MaxProb |
|---|---|---|
| 84,74 | 84,52 | 82,31 |

**Table 6.13:** Averages of the output level methods

### 6.2.4  Conclusion RQ2

With the accumulated knowledge on the design levels, it is now possible to answer RQ2. It can be seen that there is no such thing as the one true ensemble method. Many methods only bring added value under certain conditions. However, it became clear that Partitioning, Charvote, WeightedCharVote and AvgProb do not bring any recommendable added value. On the one hand, it became clear that on dataset level CompleteData as well as KFOLD, and on outpute level WordVote as well as MaxProb delivered the best results. On the other hand, Baggigng and WeightedWordVoting were also able to achieve acceptable Word Accuracies, but they rarely reached the maximum. At base learner level, it was found out that neither the heterogenous approach nor the homogenous approach were better than the other and that both can be useful. It also became clear that SimpleHTR is not much useful as a base learner. In contrast, both TrOCR and AttentionHTR scored well, with TrOCR being more suitable for the Duke Dataset and AttentionHTR more suitable for the IAM

Dataset. All these statements are of course only to be understood by looking at the decisions made before.

## 6.3 Research Question 3

Finally, to RQ3: "Can Ensemble Learning add significantly better value on smaller datasets? " The idea to answer this question was to compare the values of the small Duke Dataset with the large IAM Dataset. First of all, it can be said that the accuracy values on the Duke Dataset are significantly better than on the IAM Dataset. This is proven for example by the averages of the Word Accuracy of the ensemble outputs. The average of all Duke values of the ensemble outputs is 81.47% compared to the IAM average of 80.13%. Without the WordAccuracies of the Partitioning method with its strong outlayers it is even 92,88% against 81,56.%. But the Duke Dataset is smaller with the amount of test data and the character list of the IAM Dataset is much larger. Therefore, the absolute values cannot be used for the comparison. For this reason, the values of the difference of the respective highest base learners to the most accurate output level method will be used, which can be seen in Table 6.14. In the table it can be noticed that 2 times an ensemble of the Duke Dataset reached a larger distance to the base learner and 2 times an ensemble of the IAM Dataset. Also looking at the numbers, it is noticeable that they do not differ much. The largest gap between two differences is just about 3 percentage points. With so few numbers and so few differences, definitely no side can be favored here. It is also already indirectly known from RQ1 that ensemble learning added value for both datasets. Therefore, it can be concluded for RQ3: No, Ensemble Learning could not prove a higher added value here on a smaller dataset than compared to a larger one. Nevertheless, Ensemble Learning was able to achieve a higher accuracy on both datasets. Consequently, Ensemble Learning can be used with small as well as with large datasets in order to achieve an increased word accuracy.

| OCR Modell | Best Duke Single or Base Leaerner | Best Duke Ensemble WA | Duke Difference | Best IAM Single or Base Leaerner | Best IAM Ensemble WA | IAM Difference |
|---|---|---|---|---|---|---|
| TrOCR | 96,82 | 98,09 | 1,27 | 87,43 | 85,53 | -1,9 |
| Attention | 94,07 | 96,4 | 2,33 | 87,63 | 89,84 | 2,21 |
| simple | 91,21 | 94,92 | 3,71 | 76,73 | 82,4 | 5,67 |
| All 3 Models | 96,82 | 98,09 | 1,27 | 87,63 | 90,7 | 3,07 |

**Table 6.14:** Differences of the highest base learners to the most accurate output level method

# 7 Conclusions and Outlook

For the bachelor project 2021 "Human in the Loop: Deep Learning of Handwritten Patient Records" of Professor Lippert's research group, Ensemble Learning was looked at in order to achieve improved accuracy for the digitization of patient records.

The main question of this work was to find out with RQ1:"Can Ensemble Learning improve the Accuracy of modern OCR methods?". The answer to this question is clearly YES. Ensemble Learning offers itself, if one can overlook the disadvantages, as a very good tool to bring out some last percentage points in the accuracy of OCR. Here, however, the subliminal non-measurable, but also theoretical, advantages of Ensemble Learning such as better generalization and protection against overfitting are particularly worth mentioning. Therefore Ensemble Learning can help for the digitization of the above mentioned records.

In this context with RQ2, it was then clarified which ensemble learning methods add the most value. The main finding here was that there is not the one true ensemble method. Multiple ensemble methods can add more or less value given different circumstances. However, the most promising results here were achieved at the dataset level by CompleteData and KFOLD , at the BaseLearner level by TrOCR for Duke and by AttentionHTR for IAM, and at the output level by WordVoting and MaxProb. If these methods will be used in a new OCR project, it is very well possible that they could provide an improved accuracy. Also some here not usable methods were identified. These were Partitioning, CharVoting, WeightedCharVoting and AvgProb. The remaining methods were still shown to be relevant, although they did not produce the best results.

Since the dataset used for the Bachelor project mentioned in the introduction was very small, the last question to be answered with RQ3 was whether Ensemble Learning can help more on small datasets. Here it was found out that it makes no difference whether the dataset is large or small. Ensemble Learning added value for both datasets.

Overall, the most important finding of this work is that once again the potential of Ensemble Learning has been revealed. Consequently, ensemble learning as a research question is still relevant for OCR and can offer decisive added value. In future work, the methods looked at above could be looked at again with a different experimental setup (initializations), different datasets, or different OCR models.

# Chapter 7 Conclusions and Outlook

The question of the optimal number of base learners could even be evaluated in a separate paper. Also the approaches mentioned in the chapter ineligible methods can be used for further research. Stacking, Boosting or SnapshotEnsembling could be particularly interesting here. The topic EnsembleLearning in combination with OCR remains promising.

# Appendix

|       | TrOCR | AttentionHTR | SimpleHTR |
|-------|-------|--------------|-----------|
| Duke  | 1.39  | 2.40         | 5.72      |
| IAM   | 18.04 | 7.05         | 10.67     |

**Table 7.1:** Character Error Rates of Single OCR Modells

| Dataset Methods | Base Learners | | | | | OutputCombination | | | | | |
|---|---|---|---|---|---|---|---|---|---|---|---|
| | 1 | 2 | 3 | 4 | 5 | WordVote | CharVote | Weighted-WordVote | Weighted-CharVote | MaxProb | AvgProb |
| Complete Data | 7,94 | 1,54 | 10,18 | 9,24 | 1,14 | 2,90 | 2,44 | 1,73 | 5,13 | 3,63 | 6,57 |
| Bagging | 3,68 | 3,92 | 1,94 | 2,03 | 1,54 | 1,27 | 3,41 | 1,50 | 3,44 | 2,16 | 3,09 |
| K-FoldCross | 1,61 | 1,88 | 1,11 | 10,50 | 1,11 | 0,60 | 1,62 | 0,65 | 1,74 | 2,45 | 3,82 |
| Partitioning | 131,31 | 114,91 | 93,73 | 110,16 | 101,81 | 114,91 | 134,89 | 114,91 | 134,42 | 122,64 | 130,42 |

**Table 7.2:** Character Error Rates of homogenous TrOCR Ensembles on Duke Data

| Dataset Methods | Base Learners | | | | | OutputCombination | | | | | |
|---|---|---|---|---|---|---|---|---|---|---|---|
| | 1 | 2 | 3 | 4 | 5 | WordVote | CharVote | Weighted-WordVote | Weighted-CharVote | MaxProb | AvgProb |
| Complete Data | 7,18 | 14,32 | 16,13 | 14,86 | 12,11 | 10,16 | 15,95 | 10,16 | 17,90 | 9,51 | 11,44 |
| Bagging | 12,78 | 13,19 | 23,26 | 20,03 | 19,20 | 12,52 | 24,40 | 12,52 | 26,25 | 14,02 | 17,16 |
| K-FoldCross | 13,25 | 15,23 | 11,73 | 20,68 | 19,75 | 12,59 | 22,30 | 12,59 | 23,13 | 13,77 | 15,54 |
| Partitioning | 14,89 | 11,16 | 43,03 | 12,36 | 12,88 | 11,36 | 26,81 | 11,36 | 28,28 | 12,75 | 15,11 |

**Table 7.3:** Character Error Rates of homogenous TrOCR Ensembles on IAM Data

| Dataset Methods | Base Learners | | | | | OutputCombination | | | | | |
|---|---|---|---|---|---|---|---|---|---|---|---|
| | 1 | 2 | 3 | 4 | 5 | WordVote | CharVote | Weighted-WordVote | Weighted-CharVote | MaxProb | AvgProb |
| Complete Data | 1,99 | 2,88 | 2,55 | 2,37 | 3,02 | 1,92 | 2,69 | 1,97 | 2,69 | 1,24 | 1,57 |
| Bagging | 3,87 | 3,15 | 3,19 | 2,59 | 3,87 | 1,98 | 3,68 | 2,58 | 3,78 | 2,23 | 2,68 |
| K-FoldCross | 2,61 | 2,65 | 3,49 | 2,49 | 3,16 | 1,67 | 2,82 | 1,79 | 3,00 | 1,20 | 1,86 |
| Partitioning | 13,92 | 12,10 | 11,17 | 13,11 | 11,03 | 8,51 | 2,28 | 8,73 | 12,93 | 7,70 | 9,46 |

**Table 7.4:** Character Error Rates of homogenous AttentionHTR Ensembles on Duke Data

| Dataset Methods | Base Learners | | | | | OutputCombination | | | | | |
|---|---|---|---|---|---|---|---|---|---|---|---|
| | 1 | 2 | 3 | 4 | 5 | WordVote | CharVote | Weighted-WordVote | Weighted-CharVote | MaxProb | AvgProb |
| Complete Data | 7,03 | 7,19 | 6,75 | 7,03 | 7,25 | 6,01 | 7,43 | 6,03 | 7,52 | 5,82 | 6,01 |
| Bagging | 7,78 | 7,86 | 7,11 | 7,38 | 7,79 | 6,47 | 8,09 | 6,51 | 8,31 | 6,24 | 6,33 |
| K-FoldCross | 7,28 | 7,43 | 7,11 | 7,39 | 7,15 | 6,21 | 7,57 | 6,43 | 7,75 | 6,00 | 6,14 |
| Partitioning | 8,69 | 8,51 | 9,07 | 8,62 | 8,74 | 7,34 | 9,28 | 7,47 | 9,45 | 6,95 | 7,32 |

**Table 7.5:** Character Error Rates of homogenous AttentionHTR Ensembles on IAM Data

| Dataset Methods | Base Learners | | | | | OutputCombination | | | | | |
|---|---|---|---|---|---|---|---|---|---|---|---|
| | 1 | 2 | 3 | 4 | 5 | WordVote | CharVote | Weighted-WordVote | Weighted-CharVote | MaxProb | AvgProb |
| Complete Data | 5,65 | 5,91 | 5,42 | 4,39 | 6,29 | 3,13 | 6,16 | 3,42 | 3,92 | 3,44 | 4,66 |
| Bagging | 5,27 | 7,60 | 4,44 | 5,56 | 5,19 | 3,81 | 5,87 | 3,79 | 4,54 | 3,53 | 4,28 |
| K-FoldCross | 3,19 | 3,51 | 3,17 | 3,80 | 3,17 | 2,16 | 3,53 | 2,22 | 3,65 | 1,92 | 2,90 |
| Partitioning | 16,04 | 21,16 | 14,78 | 19,17 | 20,71 | 12,53 | 16,46 | 12,49 | 15,43 | 13,39 | 15,43 |

**Table 7.6:** Character Error Rates of homogenous SimpleHTR Ensembles on Duke Data

| Dataset Methods | Base Learners | | | | | OutputCombination | | | | | |
|---|---|---|---|---|---|---|---|---|---|---|---|
| | 1 | 2 | 3 | 4 | 5 | WordVote | CharVote | Weighted-WordVote | Weighted-CharVote | MaxProb | AvgProb |
| Complete Data | 11,04 | 11,47 | 11,26 | 11,38 | 11,35 | 9,02 | 12,85 | 9,47 | 13,12 | 8,70 | 9,92 |
| Bagging | 11,16 | 10,87 | 11,49 | 11,21 | 11,52 | 8,83 | 12,62 | 9,26 | 13,03 | 8,72 | 9,66 |
| K-FoldCross | 10,27 | 10,13 | 10,23 | 10,35 | 10,07 | 8,21 | 11,51 | 8,62 | 11,64 | 7,95 | 8,75 |
| Partitioning | 15,97 | 16,26 | 16,01 | 15,55 | 15,62 | 12,55 | 18,30 | 12,87 | 18,83 | 11,78 | 13,44 |

**Table 7.7:** Character Error Rates of homogenous SimpleHTR Ensembles on IAM Data

| Dataset Methods | OutputCombination | | | | | |
|---|---|---|---|---|---|---|
| | WordVote | CharVote | Weighted-WordVote | Weighted-CharVote | MaxProb | AvgProb |
| Complete Data | 1,02 | 3,55 | 0,63 | 3,95 | 3,45 | 7,13 |
| Bagging | 1,35 | 5,61 | 1,19 | 6,23 | 2,34 | 4,53 |
| K-FoldCross | 1,20 | 2,95 | 1,68 | 2,89 | 1,26 | 2,80 |
| Partitioning | 8,89 | 51,16 | 8,21 | 59,55 | 79,41 | 90,06 |

**Table 7.8:** Character Error Rates of heterogenous Ensembles on Duke Data

| Dataset Methods | OutputCombination | | | | | |
|---|---|---|---|---|---|---|
| | WordVote | CharVote | Weighted-WordVote | Weighted-CharVote | MaxProb | AvgProb |
| Complete Data | 5,63 | 11,98 | 5,86 | 13,38 | 6,97 | 10,30 |
| Bagging | 5,96 | 16,33 | 6,13 | 18,74 | 8,48 | 12,89 |
| K-FoldCross | 5,98 | 11,96 | 5,83 | 14,49 | 7,33 | 10,32 |
| Partitioning | 6,95 | 25,72 | 7,34 | 27,43 | 9,87 | 15,26 |

**Table 7.9:** Character Error Rates of heterogenous Ensembles on IAM Data

# Declaration of Authorship

I hereby declare that this thesis is my own unaided work. All direct or indirect sources used are acknowledged as references.

Potsdam, July 29., 2022   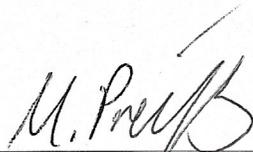
Martin Preiß